\documentclass[11pt]{article}
\usepackage[T1]{fontenc}\usepackage[utf8]{inputenc}
\usepackage[margin=1in]{geometry}
\usepackage{booktabs,graphicx,amsmath,microtype,placeins}
\usepackage[font=small,labelfont=bf,skip=6pt]{caption}
\usepackage[hidelinks]{hyperref}

\title{CulturalCrosswords:\\A Multilingual Benchmark of Answer Constraints\\and Grid Completion}
\author{\begin{tabular}{c}
Isaac Saxton-Knight \quad Gabriel Saxton-Knight \quad Diego Oliveira\\[3pt]
Tidor-Vlad Pricope \quad Aston Garratt \quad Jocelyn D'Arcy\\[6pt]
{\small Poindexter Labs}
\end{tabular}}
\date{}
\hypersetup{pdfauthor={Isaac Saxton-Knight, Gabriel Saxton-Knight, Diego Oliveira, Tidor-Vlad Pricope, Aston Garratt, Jocelyn D'Arcy}}
\begin{document}\maketitle
\begin{abstract}
When several answers fit a riddle, exact-match scoring can penalise a valid response. CulturalCrosswords tests how answer constraints affect large language models (LLMs) solving cultural riddles, whose solutions draw on knowledge specific to a community. We constructed 53 crosswords from crowdsourced riddles in 42 languages and evaluated 11 LLM configurations on five versions of each clue set: plain batches, batches with lengths, batches with lengths and partial-answer hints, and crosswords requesting an answer list or a completed grid. On 962 clues with alternative answers, lengths improve agreement with the reference by 22.8 percentage points, compared with 8.4 points when accepted alternatives also receive credit. Part of the gain therefore appears to come from selecting a required answer form. Crossword input increases the share of all crossings with compatible answers by 9.7 points; requiring a completed grid reveals large differences between models. The benchmark distinguishes agreement with a reference from the ability to produce answers that fit together and occupy the correct cells.
\end{abstract}
\section{Introduction}
A \emph{cultural riddle} asks the solver to identify something from clues that draw on knowledge and language specific to a community, for example a local custom or a nickname for a landmark. CulturalRiddles uses such riddles to test whether large language models (LLMs) can apply cultural knowledge to interpret clues~\cite{nguyen2026culturalriddles}. Several names or interpretations may fit a clue. Comparing a response with a single reference answer can therefore mark a valid solution as incorrect.

Constraints can narrow the possible answers. A specified length may distinguish two names for the same thing, while supplied letters exclude further candidates. In a crossword, \emph{crossings} are cells shared by two answers: solving one clue constrains what can fit the other. We distinguish \emph{answer-form selection}, which selects a particular wording, from \emph{semantic disambiguation}, which distinguishes meanings. Either can improve an exact-match score. Crossword constraints also impose demands on the model, since its answers must agree at shared cells and occupy the correct positions.

We introduce \textbf{CulturalCrosswords}, a benchmark of 53 crosswords constructed from CulturalRiddles, covering 42 languages and 42 countries. To study these effects, we hold the clues fixed and vary the information supplied and the response required. Each clue set appears as a plain batch (all clues in one prompt), a batch with lengths, a batch with lengths and partial-answer hints, and crossword input requesting either an answer list or a completed grid. We evaluate 11 LLM configurations to address four research questions:
\begin{enumerate}\setlength{\itemsep}{0pt}
\renewcommand{\labelenumi}{\textbf{RQ\arabic{enumi}.}}
\item\label{rq:selection} Do lengths and hints improve agreement with the reference answer, agreement that also credits accepted alternatives, or both?
\item\label{rq:compatibility} Does crossword structure help models produce mutually consistent answers beyond what lengths alone provide?
\item\label{rq:placement} How well can models return those answers in the correct grid positions?
\item\label{rq:benchmark} How does performance vary across models, task formats and language--country puzzle sets?
\end{enumerate}

The benchmark allows these comparisons on the same clues across formats. Lengths produce substantially larger gains under reference-only scoring than when alternatives receive credit, suggesting that some apparent improvement concerns answer form. Crossword input improves compatibility between answers, yet some models that supply accurate answer lists struggle to fill the grid. Scores across these tasks show where model performance differs.

\subsection{Related work}
The CulturalRiddles companion study accepts valid alternatives when evaluating open-ended answers, using rule-based checks, native-speaker annotation and model judges~\cite{nguyen2026culturalriddles}. BLEnD examines cultural knowledge across languages and regions~\cite{myung2024}; AmbigQA elicits plausible answers and disambiguating rewrites~\cite{min2020}. Our comparison applies constraints to fixed clue sets and scores the outputs both against the reference and against a set that includes accepted alternatives.

Down and Across provides crossword puzzles and clue--answer pairs with complementary evaluation measures~\cite{kulshreshtha2022}. CrossWordBench evaluates English and Chinese puzzles, including puzzles adapted from question-answer datasets, and studies supplied letters and crossing constraints~\cite{leng2025}. Buddhi-Pragati provides generated puzzles in 19 language subsets, prioritising Indian cultural content~\cite{buddhipragati}. To our knowledge, CulturalCrosswords is the first crossword benchmark for LLM evaluation to cover more than 40 languages. Its grid representation follows CrossWordBench, with rows and columns of text cells. Whereas Saha et al.\ combine language-model clue solving with search for complete grids~\cite{saha2025}, we score the answers and grids returned by the models without an external solver or repair from the reference solution.

\section{The CulturalCrosswords benchmark}\label{sec:construction}
\subsection{Source and construction}
The crosswords use crowdsourced CulturalRiddles data supplied by Cohere Labs' open science collaborators~\cite{nguyen2026culturalriddles}.\footnote{Public dataset: \url{https://huggingface.co/datasets/CohereLabsCommunity/CulturalRiddles}; attribution and evaluation-only terms: \url{https://github.com/Cohere-Labs-Community/CulturalRiddles/blob/main/data/LICENSE}. Our crosswords were constructed from the data supplied before this public release.} Each puzzle represents a language--country pair, such as English in Canada. The 53 puzzles contain 5,057 entries (individual Across or Down answers) and 5,047 distinct clue--reference pairs, where the reference is the answer in the saved solution. Ten pairs repeat within puzzles; none repeat across puzzles. These selected sets are not representative samples of languages or cultures.

We built the puzzles in Crossword Compiler 11.51\footnote{\url{https://www.crossword-compiler.com/history.html}}, initially aiming to include all 100 riddles in each source set. Entries were excluded if their answers could not fit within a $50\times50$ grid or connect to another answer through a shared cell; clues judged uninterpretable were also excluded. Some language sets could not be used because the answers lacked feasible crossings, their writing units could not be segmented with the available procedure, or the source data were judged unreliable. Appendix~\ref{app:construction} details these construction decisions and the source matching. All conditions preserve the exact crossword clues and use the saved solutions for the required answers, lengths and positions. Across and Down entries remain distinct even when they share a number.

\subsection{Five matched conditions}
The five tasks use the same clues for a given puzzle, but differ in input length and information content (Table~\ref{tab:conditions}). A, B, C and E request a list pairing each answer with its clue identifier; D requests a completed grid. Appendix~\ref{app:prompts} gives prompt excerpts and describes the complete archived prompts. Lengths and hints count \emph{written units}: the content of one stored grid cell, which may contain multiple encoded characters, such as a base letter and combining marks. We denote the length of entry $i$ by $\ell_i$.

\begin{table}[htbp]\centering\small
\caption{Task conditions. Plain batches retain the shared language and answer conventions; they omit lengths, hints and crossings.}\label{tab:conditions}
\begin{tabular}{clp{8.1cm}}\toprule Condition & Short name & Information and required response\\\midrule
A & Plain & Full clue batch; answer list.\\
B & Length & A plus required answer lengths; answer list.\\
C & Hints & B plus correct units at one quarter of each answer's positions, rounded down ($\lfloor\ell_i/4\rfloor$); answer list.\\
D & Grid & Blank grid and clues with start coordinates and directions; completed grid. Lengths are inferred from the grid.\\
E & Crossword/list & Blank grid, entry paths, explicit lengths and clues; answer list.\\\bottomrule
\end{tabular}\end{table}

Hint positions are selected once using the rule in Appendix~\ref{app:construction} and held fixed across models. Because hints supply reference units directly whereas crossings link answers that must still be inferred, C and E provide different kinds and amounts of information.

\subsection{Grid representation}
\begin{figure}[htbp]\centering
\includegraphics[width=\textwidth]{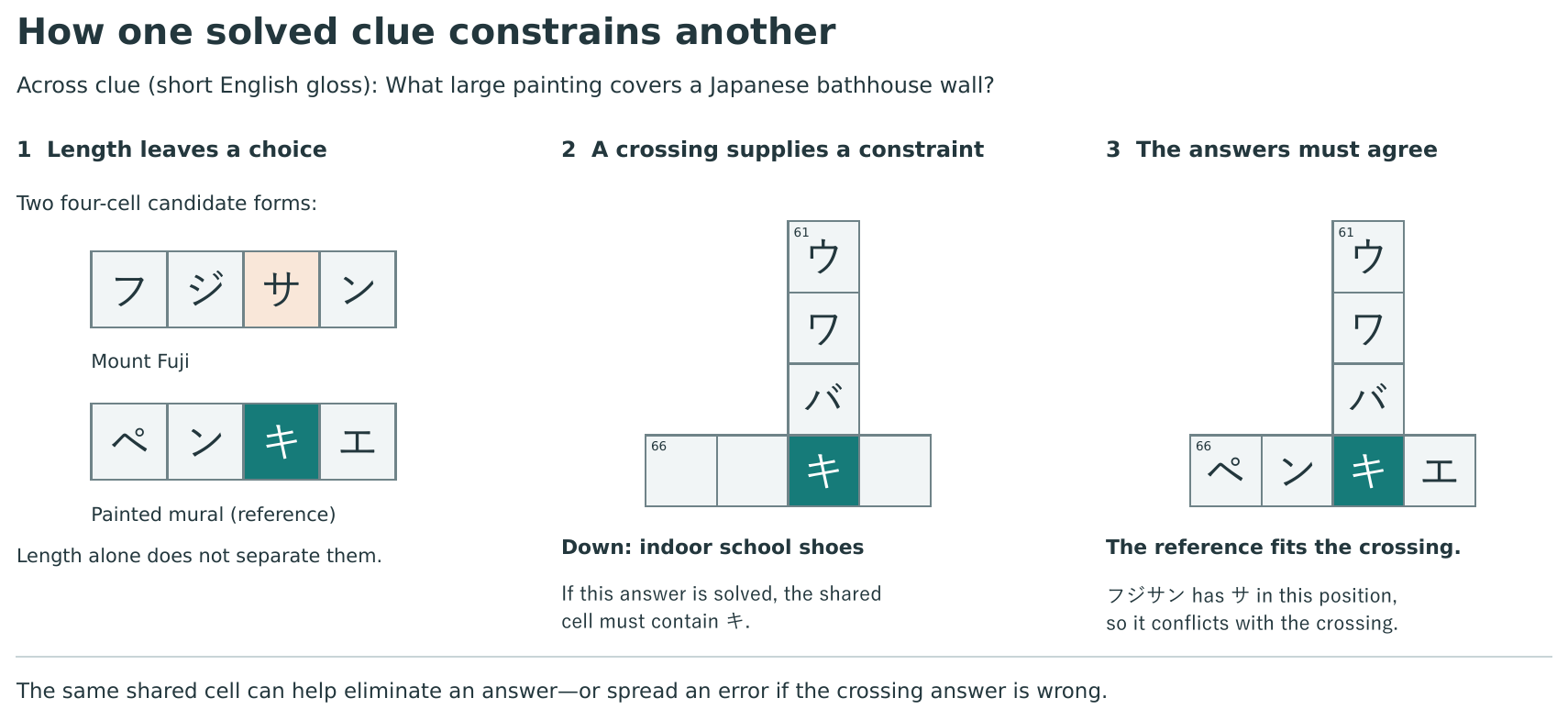}
\caption{A Japanese crossing. Both candidates occupy four katakana cells. Solving the school-shoes clue fixes the third Across unit and excludes the Mount Fuji form. Relative source-grid positions are preserved; clues are shortened English glosses. This illustrates compatibility given a correct crossing, rather than observed model reasoning. An incorrect crossing can propagate an error.}\label{fig:crossing-example}
\end{figure}

We use a text-grid representation, as in CrossWordBench~\cite{leng2025}. Let $G$ be a grid with $H$ rows and $W$ columns of blocked or fillable cells. Entry $i$ has clue $c_i$, length $\ell_i$ and ordered cell path $p_i=(p_{i1},\ldots,p_{i\ell_i})$. Answers must agree at shared cells:
\[
p_{ik}=p_{jt}\quad\Longrightarrow\quad a_{ik}=a_{jt},
\]
where $a_{ik}$ is the $k$th written unit of answer $i$. We preserve whole-cell units and logical answer order, including right-to-left entries. We group puzzles by their answer script, following Unicode writing-system terminology~\cite{unicodeScripts}: 32 use alphabets, nine use abjads (consonant-centred scripts), 11 use abugidas (consonants with vowel modifications), and one uses Japanese katakana, a syllabary (symbols for sound units). These groups describe the stored answers, not every script used in a language. Figure~\ref{fig:crossing-example} illustrates the information a crossing can provide.

\section{Experimental setup and evaluation}\label{sec:execution}
\subsection{Models and execution}
We evaluated 11 LLM configurations, each comprising a named model and its recorded settings, on all 53 puzzles and five conditions. One response per combination enters the analysis, giving 2,915 responses. Technical failures were retried, as were selected incomplete answer lists, for which we added instructions to return JSON (a machine-readable text format) directly in the response. Incorrect answers did not trigger retries. The recovery procedure is described below. A retained response may combine continuation requests asking the model to finish an interrupted output.

The models are GPT-6 Astra (Astra), GPT-6.1 Sol (Sol), GPT-5.5, Gemini 4 Argon (Argon), Gemini 3.8 Flash, Gemini 3.7 Flash, Gemini 3.1 Pro, Claude Fable 5.1, and Claude Opus 5.5, 4.8 and 4.7. The shared configuration requests high reasoning and one result, with code execution and web search disabled. All response records report no tool access or use. The implementation sets temperature (sampling randomness) to 0 for Gemini 3.1 Pro and omits it for the other ten models. Top-$p$, the probability cutoff for token sampling, and the random seed are omitted throughout. Appendix~\ref{app:models} lists model identifiers and the evidence for these settings.

Generation limits and continuation policies differ across providers. D uses crossword-specific system instructions, whereas A, B, C and E use shared instructions for general science, technology, engineering and mathematics problem solving. Comparisons measure performance under these execution policies without holding computation budgets constant. Appendix~\ref{app:settings} documents the limits, termination records and token usage.

\subsection{Answer agreement, coverage and placement}\label{sec:evaluation}
The primary metric, \emph{reference-answer agreement}, is the proportion of expected clues whose extracted answers match the saved solution. Before matching, we apply \emph{normalisation} according to the prompt's rules for casing, spacing and written units. Specified formatting differences are allowed; synonyms and missing content are not added. Agreement with the required answer form does not establish whether other interpretations also fit the clue.

For puzzle $p$ with $N_p$ entries, let $y_{pmti}=1$ for a reference match and $f_{pmti}=1$ for a fully specified answer (zero otherwise) for model $m$, condition $t$ and entry $i$. Agreement and \emph{entry coverage} are
\[
A_{pmt}=\frac{1}{N_p}\sum_i y_{pmti},\qquad
C_{pmt}=\frac{1}{N_p}\sum_i f_{pmti}.
\]
Missing, unknown, conflicting and incomplete answers score zero agreement; correct entries in partial responses retain credit. A fully specified but wrong answer contributes to coverage. Agreement among fully specified answers is a secondary measure that excludes missing answers from its denominator; it is undefined when no complete answers are supplied.

Answer extraction removes unambiguous formatting and uses explicit identifiers to link text to clues, without supplying missing answers or choosing among unresolved alternatives. In D, each complete answer must also occupy the correct coordinates. Correctly placed entries can receive credit in an otherwise malformed grid, but we do not shift rows, pad cells or repair crossings. \emph{Structural validity}, reported separately, checks dimensions, cell-unit representation and blocked-cell layout rather than answer correctness. Extraction audits and normalisation checks are documented in Appendix~\ref{app:recovery}.

\subsection{Incomplete responses and alternative answers}
Thirty original responses containing only an ellipsis (\texttt{...}) remain nonanswers. A separate set of 322 incomplete Claude answer-list records was selected for recovery: 145 Opus 5.5, 127 Opus 4.7 and 50 Fable 5.1. These texts often referred to answers absent from the saved response. They do not establish whether the model omitted those answers or whether capture was incomplete. We replaced all 322 records with usable responses obtained under explicit inline-output instructions, preserving the clues and their constraints, reference answers and intended model settings. Blank or incorrect answers in otherwise usable responses were retained. Original results and selection flags are preserved for sensitivity analysis. Appendix~\ref{app:recovery} specifies selection and extraction.

For clues with alternative answers, \emph{reference-or-alternative agreement} credits either the reference answer or a listed alternative. The list is incomplete, so unlisted answers may also be correct. Compliance with lengths and hints is checked separately; an acceptable alternative can still violate those constraints.

\subsection{Crossing consistency}
\emph{Crossing consistency} measures whether answers give the same written unit at a shared cell. For B and E, we map complete answers onto the saved grid. A crossing is checkable when both answers have the required length and map unambiguously under the unit and direction rules. For $T$ total crossings, $K$ checkable crossings and $M$ agreeing crossings, we report $K/T$ (checkable-crossing coverage), $M/K$ (conditional consistency), and $M/T$ (consistent crossings out of all crossings). Conditional consistency is undefined when $K=0$. Compatible answers can still be wrong.

\subsection{Statistical analysis}\label{sec:analysis}
We compare matched responses to measure the addition of lengths (B--A), partial-answer hints (C--B), and crossword information with answer-list output (E--B). E--A describes the combined change. E--D compares whole tasks because instructions, clue IDs, explicit lengths and execution policies also differ; this contrast cannot isolate output format.

We weight puzzles equally, preventing larger clue sets from dominating:
\[
\overline A_{mt}=\frac{1}{53}\sum_{p=1}^{53}A_{pmt},\qquad
\Delta_{m,t-s}=\frac{1}{|P|}\sum_{p\in P}(A_{pmt}-A_{pms}),
\]
where $P$ is the matched puzzle set. We also report summaries weighting each clue equally, or averaging country variants within a language before weighting languages equally. Aggregates across models average the fixed panel within each puzzle, then weight puzzles equally.

Uncertainty estimates preserve paired observations~\cite{dror2018}. We use a \emph{language-cluster bootstrap}: 10,000 samples of whole language groups drawn with replacement, keeping country variants together and puzzles equally weighted. The resulting 95\% intervals describe variation across the collection. They do not estimate repeated-generation uncertainty and are unadjusted for multiple comparisons. Puzzle-level resampling checks the clustering choice; Appendix~\ref{app:sensitivity} records seeds and comparison counts.

The primary analysis uses all 11 configurations, including recovery responses. To assess dependence on recovery, we repeat comparisons using the original outputs or only response pairs not selected for recovery. A separate eight-model analysis excludes Opus 4.7, Opus 5.5 and Fable 5.1, the models that received recovery reruns. We also omit each model in turn across all conditions, including language comparisons and regression. Selection flags continue to refer to original responses after replacement, and answer-transition counts include both gains and losses. The 278,135 entry observations reuse clues across conditions and models; they are not independent trials.

Ordinary least-squares regression relates puzzle-level mean E--B gains to clue count, mean answer length and \emph{crossing density}: the fraction of positions shared with another entry, averaged over entries. The 53 puzzles receive equal weight; intervals use 10,000 language-cluster draws~\cite{cameron2015} (Appendix~\ref{app:regression}).

\section{Results}\label{sec:results}
\subsection{Answer forms and disambiguation}\label{sec:disambiguation}
The alternative-answer analysis (RQ\ref{rq:selection}) covers 1,890 forms for 962 clues in 38 puzzles and 30 languages. We score each response twice: against the crossword's reference alone, and against the reference plus listed alternatives. The list is incomplete and contains no Japanese entries; matching rules appear in Appendix~\ref{app:alternatives}.

\begin{figure}[tb]\centering
\includegraphics[width=\textwidth]{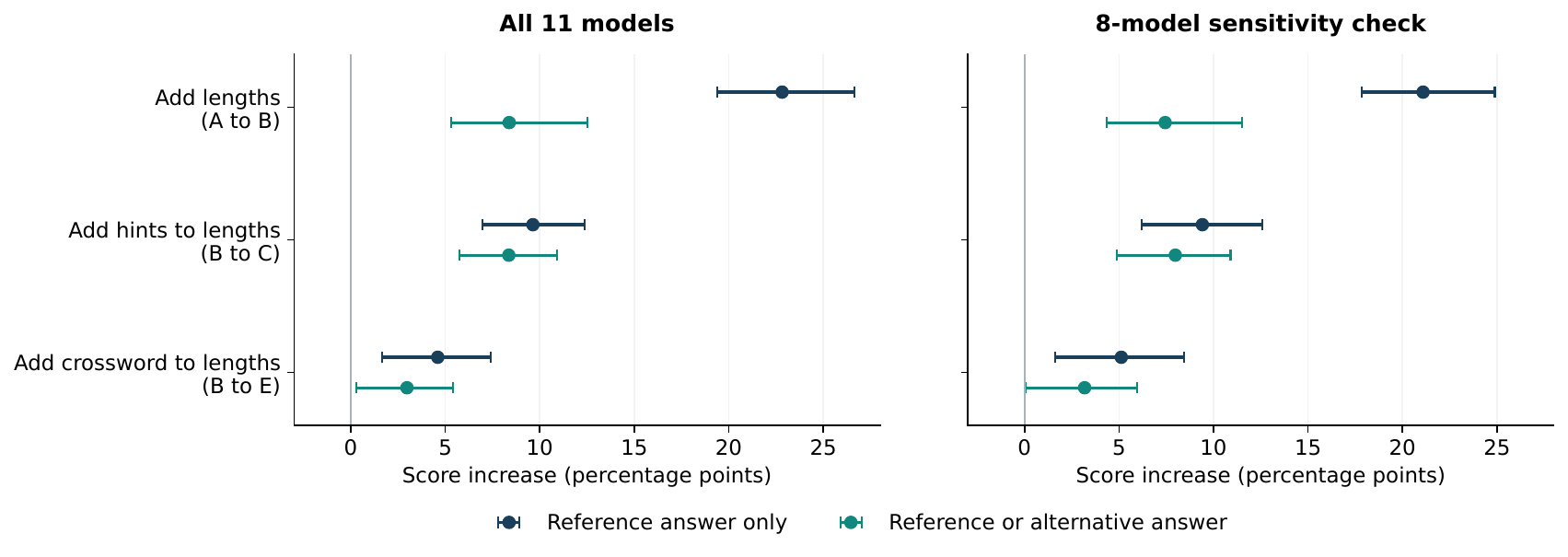}
\caption{Score increases on the same 962 clues. Blue credits reference answers only; green also credits listed alternatives. Each row adds information while retaining answer-list output. Models are averaged within each of 38 equally weighted puzzles; bars show 95\% language-cluster intervals. The right panel excludes Opus 4.7, Opus 5.5 and Fable 5.1, which received recovery reruns.}\label{fig:disambiguation}
\end{figure}

Adding lengths improves the reference-only score by 22.8 percentage points, compared with 8.4 points when alternatives also receive credit (Figure~\ref{fig:disambiguation}). Part of the gain therefore appears to come from choosing the required answer form. GPT-5.5, for example, changes from \texttt{COCACOLA} to \texttt{COKE} after receiving lengths, substituting one name for the same drink. The change from \texttt{SHARECROPPING} to \texttt{JIMCROWLAWS} on another clue may instead reflect a different interpretation (Appendix~\ref{app:alternatives}).

Hints added to lengths improve the two scores by 9.6 and 8.4 points; adding crossword information improves them by 4.6 and 3.0 points. The eight models without recovery reruns show the same pattern. These gains extend to scoring that accepts alternatives, and every comparison requests an answer list, without requiring grid placement. Neither score, nor the difference between them, directly measures semantic disambiguation because the alternative list is incomplete.

\subsection{Crossword structure improves answer compatibility}
Crossword input improves the compatibility of returned answers (RQ\ref{rq:compatibility}). Averaged across the evaluated models, the share of all crossings with compatible answers rises from 61.8\% with lengths alone (B) to 71.5\% with crossword input (E), despite lower checkable-crossing coverage (Table~\ref{tab:crossing-consistency}). The +9.7-point change has an interval of [7.5, 11.9], whereas reference agreement increases by 2.8 points [1.3, 4.3]. Thus, the improvement in compatibility exceeds the gain in reference agreement, although these scores cannot establish how the models used crossings during reasoning.
\begin{table}[t]
\centering
\small
\begin{tabular}{lrrr}
\toprule
Measure & B & E & E--B [95\% CI] \\
\midrule
Reference agreement & 70.1 & 72.9 & +2.8 [1.3, 4.3] \\
Checkable crossings & 81.2 & 74.8 & -6.4 [-9.0, -3.9] \\
Consistent / all crossings & 61.8 & 71.5 & +9.7 [7.5, 11.9] \\
\bottomrule
\end{tabular}
\caption{Crossing analysis across all 11 models and 53 puzzles. Values are equally weighted puzzle--model percentages; changes are percentage points. Intervals resample 42 language clusters. A crossing is checkable only when both complete answers map unambiguously to the required number of cells.}
\label{tab:crossing-consistency}
\end{table}

\begin{figure}[!htbp]\centering
\includegraphics[width=\textwidth]{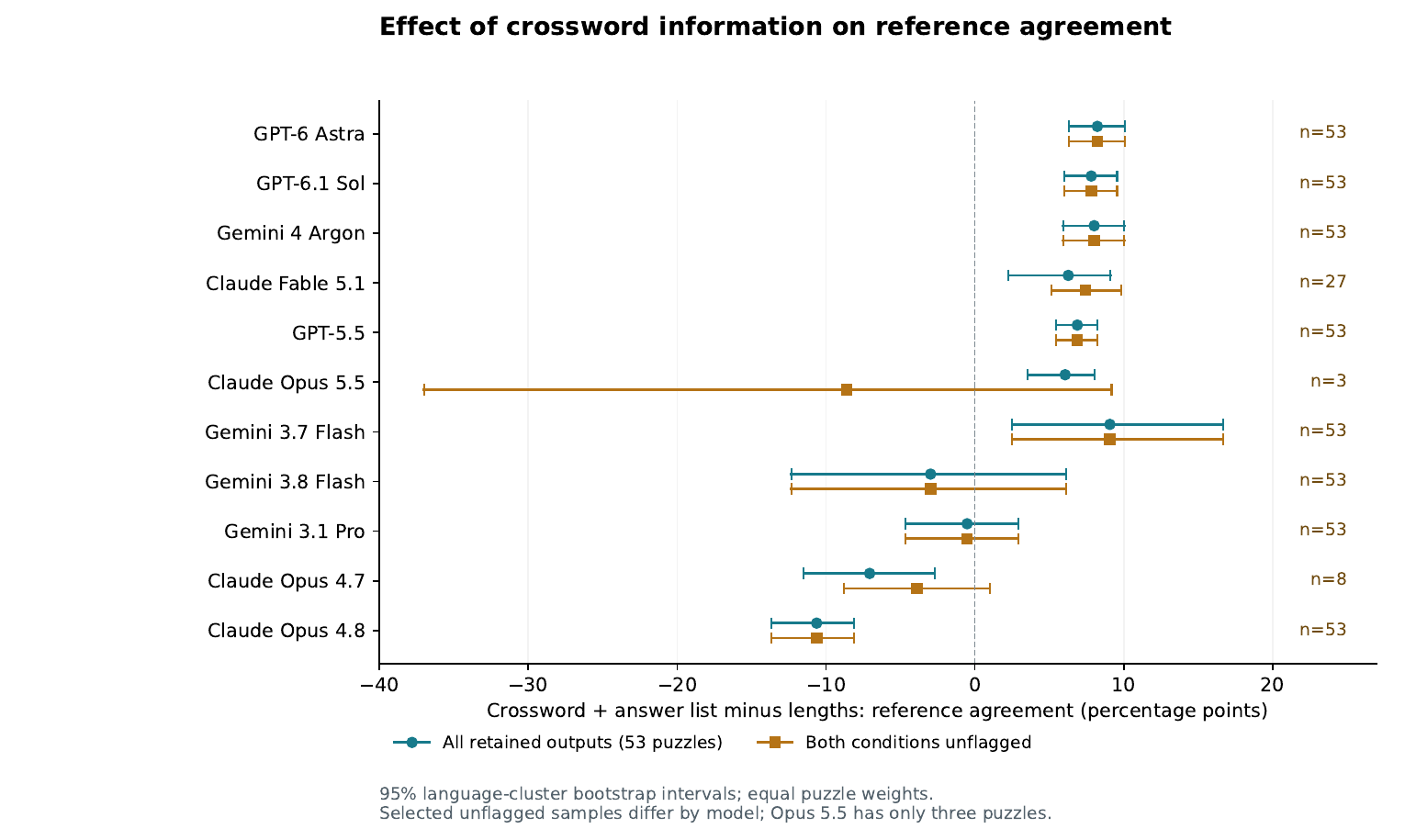}
\caption{Reference-agreement change from lengths to crossword input with answer-list output (E--B), in percentage points. Teal circles use all 53 puzzles, including recoveries. Orange squares exclude puzzles with a response selected for recovery in either condition; $n$ is the number remaining. Bars show 95\% language-cluster intervals with equal puzzle weights. Very small subsets are descriptive only.}\label{fig:crossword-effects}
\end{figure}

Reference-agreement gains vary by model (Figure~\ref{fig:crossword-effects}). Among the eight models without recovery reruns, five improve by 6.9--9.1 points; Gemini 3.8 Flash, Gemini 3.1 Pro and Opus 4.8 score lower. Recovery also affects the comparison. With clearer output instructions and additional attempts, Fable 5.1's E--B difference changes from $-4.2$ to +6.3 points (Appendix~\ref{app:sensitivity}).

\subsection{Grid completion}\label{sec:grid-completion}
Requiring a filled grid reveals substantial differences in task performance (RQ\ref{rq:placement}). Gemini 3.7 Flash scores 75.6\% with crossword input and answer-list output, falling to 5.4\% when required to fill the grid; Opus 5.5 performs similarly on the two tasks after recovery, at 82.6\% for answer lists and 80.4\% for grids. The comparisons also include the instruction and execution differences described in Section~\ref{sec:analysis}, so the gaps cannot be attributed solely to placement difficulty.

Grid structure and answer placement can also diverge. Opus 5.5 produces structurally valid grids for 44 puzzles, compared with Astra's 36, yet Astra places more reference answers correctly. Opus 4.8 returned prose nonanswers on all 53 grid tasks, ending normally without recorded token-limit stops. Its zero agreement and coverage reflect noncompletion under this protocol; agreement among supplied answers is undefined. Omitting it across conditions changes the E--D gap only slightly (26.6 versus 26.8 points; Appendix~\ref{app:sensitivity}).

\subsection{Model benchmarking}\label{sec:model-benchmark}
\begin{figure}[!htbp]\centering
\includegraphics[width=\textwidth]{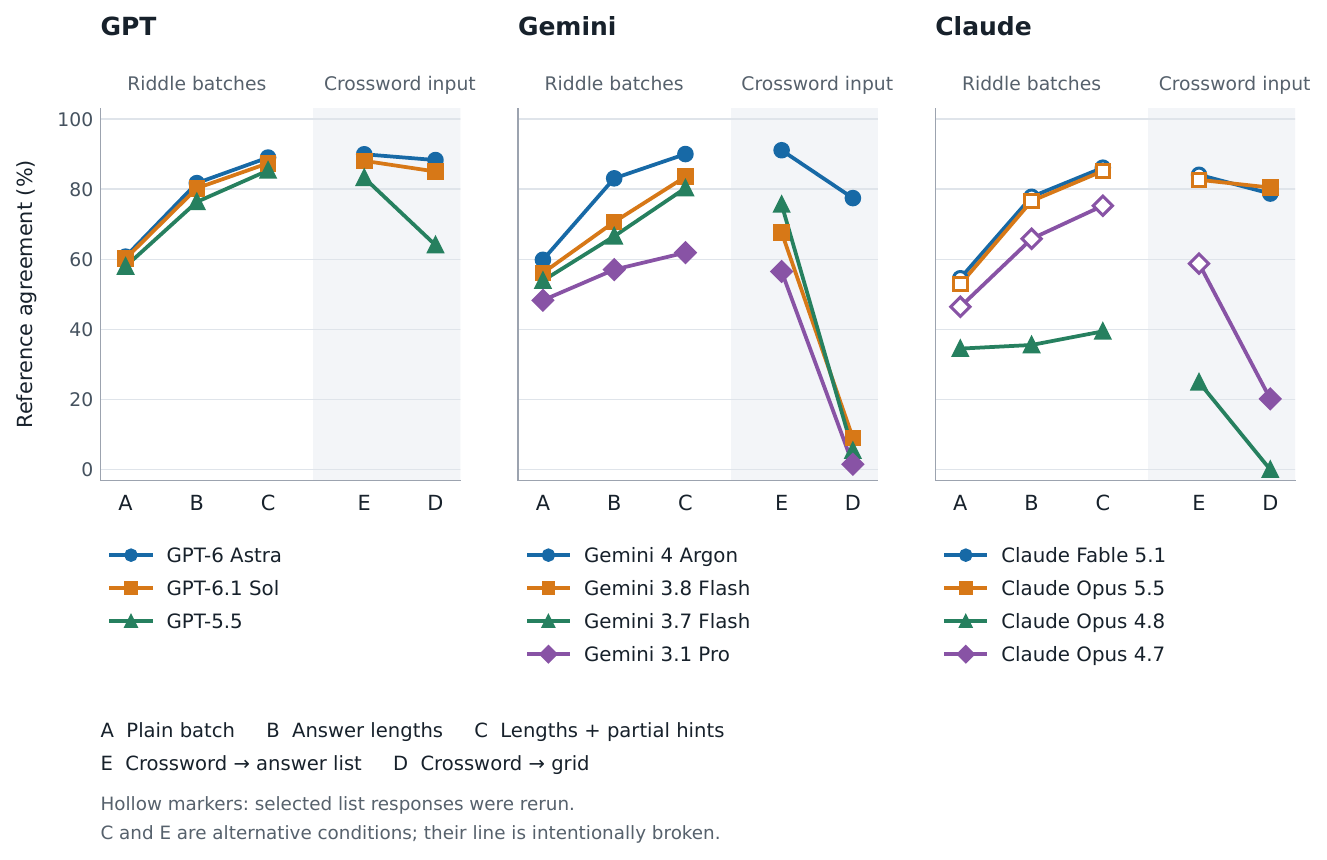}
\caption{Reference agreement, weighting 53 puzzles equally. A--B--C adds answer information; E supplies crossword structure; D requires placement. These are discrete tasks, not a continuous constraint scale. Missing answers remain in the denominator. Hollow markers identify means containing records selected for recovery.}\label{fig:profiles}
\end{figure}

The highest-scoring model depends on the task (RQ\ref{rq:benchmark}; Figure~\ref{fig:profiles}). On the same 53 puzzles, Astra has the highest observed mean for plain batches (60.7\%) and completed grids (88.3\%), while Argon leads the length (83.1\%), hint (90.0\%) and crossword/list (91.1\%) conditions. Close ranks are uncertain: Astra's 0.6-point lead over Sol in A has an interval of approximately $[-0.2,1.4]$. Appendix~\ref{app:scores} gives all scores and entry coverage, with radar profiles showing each model across the five formats (Figure~\ref{fig:model-radars}).

Mean agreement improves from A to B and from B to C for all 11 configurations, including the eight models without recovery reruns. The comparisons remain specific to the evaluated configurations and execution protocol, which includes selective recovery attempts.

\subsubsection{Variation across puzzle sets}\label{sec:variation}
\begin{figure}[htbp]\centering
\includegraphics[width=\textwidth]{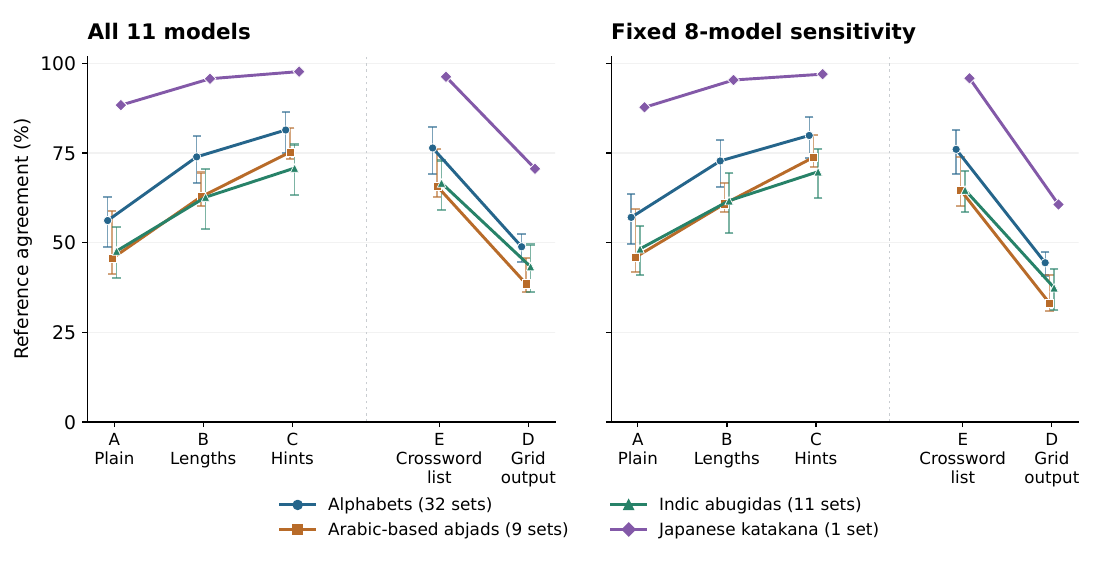}
\caption{Writing-system profiles with equal puzzle weights within groups and equal model weights within panels. Bars show language-cluster intervals. Alphabetic, abjad and abugida groups contain 27, four and ten language labels. Abjad intervals are descriptive because there are only four language clusters; Japanese has one puzzle and no population interval. D requires placement. Eight-model panels exclude Opus 4.7, Opus 5.5 and Fable 5.1, which received recovery reruns. C and E supply different information.}\label{fig:writing-systems}
\end{figure}

Performance also varies across language--country puzzle sets (RQ\ref{rq:benchmark}). Figure~\ref{fig:writing-systems} groups them by writing system and compares all models. For alphabets, Arabic-based abjads and Indic abugidas, group means increase from plain batches to length and hint conditions. Japanese scores 88.4\% in A and 97.7\% in C. Its 100 clues rule out a smaller batch as the explanation, although the short katakana answers and content of this single puzzle may contribute. High scores already occur without crossword constraints; they cannot establish general Japanese or syllabary performance.

The regression explains little variation in E--B gains. $R^2$, the fraction explained, is 0.032 overall, 0.038 for the eight-model panel and 0.017 without Opus 4.8; all three predictor intervals include zero. Answer length and crossing density are strongly correlated (Pearson $r=-0.895$), limiting their separate interpretation (Appendix~\ref{app:regression}). These results leave the differences unexplained by the fitted linear model, without establishing that puzzle structure is irrelevant.

\section{Discussion and limitations}
\paragraph{Answer selection (RQ\ref{rq:selection}).}
Lengths and hints improve both scoring measures. The much larger gain from lengths against the reference alone matters for interpretation: a model can gain an exact match by choosing another acceptable name without changing what it means. Scoring with alternatives helps identify this distinction, although the incomplete alternative list prevents us from estimating how often meaning was disambiguated. That requires native-language review of gains and losses, including answers absent from the list.

\paragraph{Crossword compatibility (RQ\ref{rq:compatibility}).}
Crossword input improves compatibility between returned answers, even where reference agreement changes modestly. Crossing checks thus capture a benefit that reference scoring alone can miss: the answers more often fit together. The returned outputs do not establish how the models used crossings during reasoning.

\paragraph{Grid completion (RQ\ref{rq:placement}).}
Success at giving answers does not guarantee success at placing them. Gemini 3.7 Flash has a large list--grid gap, whereas Opus 5.5 scores similarly on the two tasks. Differences in instructions and execution prevent attributing the gap solely to placement. Reporting answer coverage alongside agreement and correct placement shows whether a low score accompanies omissions or supplied answers; checking grid shape alone cannot establish successful solving.

\paragraph{Comparative benchmarking (RQ\ref{rq:benchmark}).}
The strongest observed model depends on the task. Astra leads plain batches and grid completion; Argon leads the other three conditions. Difficulty across language--country sets broadly aligns with the CulturalRiddles companion study~\cite{nguyen2026culturalriddles}: seven of its eight hardest sets occur in our collection, all within our 13 hardest plain-batch sets. Japanese is among the easiest in both. This agreement concerns related samples from the same corpus evaluated with different models and protocols. It is not an independent replication, and these puzzle sets cannot measure general cultural proficiency.

Applying this comparison to other short-answer benchmarks would require answer representations with enough shared written units to form connected puzzles. Whether constraints help select among acceptable answers in those datasets remains to be tested.

\paragraph{Scope and limitations.}
Language differences are confounded with cultural content and puzzle construction. Inclusion depended on answer length, shared written units, segmentation support and our judgements of clue quality, so writing-system groups differ both in clue count and in which riddles entered a grid. Equal puzzle weighting and the size-adjusted regression do not remove this selection effect. Nor do the alternative-answer list and English case review independently validate every reference.

Selected Claude responses received clearer output instructions and additional attempts, so the rankings do not compare single-attempt performance. Their improvements cannot be attributed solely to repaired capture or to the instruction change. The original-output and eight-model analyses test how findings change when recovery responses or affected models are excluded. Provider-specific generation policies and incomplete deployment metadata also prevent comparison at equal computation.

\section{Conclusion}
CulturalCrosswords tests how LLMs answer the same cultural riddles under different constraints and output requirements. Lengths and hints improve reference agreement. The smaller gains when alternatives receive credit suggest that some improvement reflects selection among acceptable answer forms. Crossword input makes returned answers more compatible, although some models still struggle to fill the grid. These differences favour reporting performance by task: an exact-match score alone cannot show whether a model supplied all answers or placed them correctly.

\paragraph{Data and code.}
The reproducibility materials contain the evaluated clues and grids, reference keys, exact task prompts, model responses, source-riddle mappings, scoring code and analysis outputs. They preserve original and recovery responses separately. The public CulturalRiddles release has attribution and evaluation-only terms; redistribution of any additional material supplied privately requires agreement with the source collaborators.

\clearpage\appendix
\section{Models and execution records}\label{app:models}\label{app:settings}
Table~\ref{tab:identifiers} lists the recorded model identifiers. These do not independently verify the vendor checkpoints used. Section~\ref{sec:execution} gives the shared settings.

\begin{table}[htbp]\centering\small
\caption{Display names and recorded identifiers.}\label{tab:identifiers}
\begin{tabular}{lp{9.5cm}}\toprule Display name & Recorded identifier\\\midrule
GPT-6 Astra (Astra) & \path{gpt-6-astra}\\
GPT-6.1 Sol (Sol) & \path{gpt-6.1-sol}\\
GPT-5.5 & \path{gpt-5.5}\\
Gemini 4 Argon (Argon) & \path{google/gemini-4-argon}\\
Gemini 3.8 Flash & \path{google/gemini-3.8-flash}\\
Gemini 3.7 Flash & \path{google/gemini-3.7-flash}\\
Gemini 3.1 Pro & \path{gemini-3.1-pro-preview-native-code-search}\\
Fable 5.1 & \path{claude-fable-5-1}\\
Opus 5.5 & \path{claude-opus-5-5}\\
Opus 4.8 & \path{claude-opus-4-8}\\
Opus 4.7 & \path{claude-opus-4-7}\\\bottomrule
\end{tabular}\end{table}

\paragraph{Configuration evidence.}
Execution metadata were matched to all 2,915 original responses by task identifier, model and condition. All records report no code or web access or use, including the model whose identifier contains ``code-search''. Reasoning metadata is \texttt{high} for all 583 D records. Of the 2,332 answer-list records, 2,164 have the label \texttt{mixed} and 168 have \texttt{high}; implementation inspection identifies \texttt{mixed} as an aggregation label rather than a provider reasoning level. Exact response-text matches to job logs corroborate requested high reasoning and disabled tools for 64 analysed responses. Some GPT response identifiers are also recorded. The configuration and execution records agree where these checks were possible, but the metadata do not identify every exact deployment or preserve every complete provider request.

\paragraph{Provider policies.}
We reconstructed request behaviour from the execution code and its version history, subject to the per-response verification limits above. Temperature is omitted except where stated; top-$p$ and seed are always omitted. Unspecified sampling parameters are left to provider defaults.
\begin{itemize}
\item \textbf{GPT-5.5, Astra and Sol:} Azure-routed Responses background calls pass configured reasoning effort and omit maximum output tokens.
\item \textbf{Opus 4.7, Opus 4.8, Opus 5.5 and Fable 5.1:} requests use the Anthropic Messages API with streaming and configured reasoning effort. High effort starts at 32,000 tokens. Eligible incomplete token-limit stops trigger continuation instructions retaining assistant content, with caps of 64,000 and then 128,000. The default permits two continuations, subject to guards and a potentially lower environment override. A usable complete structured answer may be accepted at the limit.
\item \textbf{Gemini 3.1 Pro:} native generation passes configured thinking level, sets temperature to zero, and uses structured JSON when tools are disabled. Maximum output tokens are omitted.
\item \textbf{Gemini 3.7 and 3.8 Flash:} native generation passes configured thinking level and omits maximum output tokens.
\item \textbf{Gemini 4 Argon:} background Interactions passes configured thinking level and uses structured JSON when tools are disabled. Recorded thinking/output caps are reported below.
\end{itemize}
The 4,096-token sheet field does not control native crossword execution. D uses crossword-specific system instructions; A/B/C/E use the shared technical problem-solving instructions described in Section~\ref{sec:execution}. Complete provider message sequences and exact deployment versions were not recorded in the study exports or sampled job logs.

\paragraph{Stops and usage.}
Of the 1,060 original Claude results, 442 have intermediate \texttt{max\_tokens} stops: 373 once and 69 twice, or 511 events. All finally record \texttt{end\_turn}, indicating normal termination rather than exhaustion of the token limit. Limit stops affect 145/212 D results (68.4\%) and 297/848 answer-list results (35.0\%). The records do not cover every termination or discarded attempt for other providers.

\begin{table}[htbp]\centering\small
\caption{Original Claude results with at least one intermediate limit stop, out of 53 per cell. Counts precede recovery and measure stop incidence rather than tokens consumed.}\label{tab:stops}
\begin{tabular}{lrrrrr}\toprule Model & A & B & C & E & D\\\midrule
Opus 5.5 & 0 & 5 & 16 & 43 & 51\\
Fable 5.1 & 8 & 20 & 31 & 45 & 51\\
Opus 4.7 & 6 & 38 & 46 & 39 & 43\\
Opus 4.8 & 0 & 0 & 0 & 0 & 0\\\midrule
Total / 212 & 14 & 63 & 93 & 127 & 145\\\bottomrule
\end{tabular}\end{table}

Input, output, thought and total-token usage were recorded for 50 Argon D responses, with thinking/output caps of 262,144 for 49 and 524,288 for one. The remaining three Argon D records lack caps. Token usage is not recorded for the other 2,865 responses. Caps do not measure tokens consumed. The available usage records cannot support equal-compute or general cost comparisons.

\paragraph{Recovery execution records.}
All 322 integrated recoveries have matching requested and response model identifiers and use the shared technical problem-solving wrapper. Code and web tools remain disabled, with no reported use. Of these responses, 195 have an intermediate \texttt{max\_tokens} stop; all finally record \texttt{end\_turn}. The saved recovery records report these stop histories but omit per-request sampling parameters and token usage.

\paragraph{Parser flags and noncompletion.}
All 116 response-error flags occur in D: 60 \texttt{invalid\_json} and 56 \texttt{invalid\_grid}. They comprise 53 Opus 4.8, ten Opus 4.7, 21 Gemini 3.1 Pro, and 16 each for Gemini 3.7/3.8 Flash. All 53 Opus 4.8 texts are prose noncompletion flagged as \texttt{invalid\_json}; none contains a recoverable grid answer. Each records \texttt{end\_turn} without an intermediate limit stop. Flagged responses can contain recoverable entries; unflagged responses can be incomplete. Validators differ by execution path.

\section{Construction and hint selection}\label{app:construction}
The evaluation uses 53 crossword files containing clues, reference solutions and $50\times50$ grids. The source materials include 55 cleaned workbooks (5,499 rows). The reproducibility materials preserve these inputs with source identifiers and SHA256 checksums identifying the exact file contents. An accompanying manifest records source versions and prompt revisions. Construction records describe the selection rules but do not specify individual exclusions, layout-search settings or the complete sequence of manual edits. The supplied tasks support reproduction of the evaluation, but not exact reconstruction of grid generation from the raw corpus.

\paragraph{Construction exclusions.}
We aimed to include all 100 source riddles per set. Answers were omitted if longer than a grid dimension, if the combined set could not fit, or if they could not share a written unit with another answer. We also excluded a small number of clues judged uninterpretable; this count was not recorded. Disconnected answers could include numerical strings or a different script from the rest of the set. In each of the 53 grids, all entries connect through crossings and none is isolated.

Five Chinese varieties were excluded because their answer sets did not yield sufficiently connected grids under our construction procedure. This does not imply that Chinese crosswords cannot be constructed. Kannada and Thai were omitted because the required writing-unit segmentation was not implemented. Our construction review also excluded Buginese over concerns about language and script consistency; those judgements were not independently validated. We evaluate the 53 constructed puzzles rather than all language sets in the source corpus.

The intended units were letters for alphabets and abjads, kana for Japanese, and aksharas (orthographic syllables) for abugidas. Evaluation uses the units stored in the grids. Abugida puzzles contain 72--99 entries (mean 82.7), compared with 86--100 (98.5) for alphabets, 98--100 (99.4) for abjads, and 100 for Japanese.

The 5,057 placements contain 5,047 distinct exact clue--reference pairs: ten recur within a puzzle; none across puzzles. Language--country and clue joins identify 5,037 distinct source riddle IDs: 4,208 by exact match and 829 after conservative Unicode/HTML/whitespace handling. For 20 ambiguous matches, we do not break ties using reference answers. Unresolved joins remain archived, subject to source-release terms.

\paragraph{Hint selection.}
{\emergencystretch=2em
SHA256 maps text to a fixed hexadecimal value, allowing a repeatable ordering of positions. For positions $j=1,\ldots,L$ in reading order, compute SHA256 of the UTF-8 string \texttt{seed|source\_answer\_id|j}, using seed \texttt{riddles-\allowbreak crossword-\allowbreak matched-\allowbreak v1-\allowbreak 2026-\allowbreak 10-\allowbreak 14}. Here, \texttt{source\_answer\_id} joins puzzle code, Across/Down direction and clue number with colons; it is not the corpus riddle ID. Rank positions by ascending hexadecimal digest, then ascending position to break ties. Reveal the first $\lfloor L/4\rfloor$ whole stored units in logical order, including right-to-left entries. All 5,057 hint masks (patterns of revealed and hidden positions) were reconstructed from this rule; code, positions and source hashes are archived.\par}

\section{Representations and recovery}\label{app:recovery}
C reveals 9,472 of 45,373 answer-unit occurrences (20.9\%); 369 entries shorter than four units receive no hints. Twelve puzzles use cells containing multiple Unicode code points and nine use right-to-left placement. Logical answer order is tracked separately from physical coordinates. Structural grid validity requires the expected dimensions, blocked-cell mask and one stored unit per cell, with no unreplaced period markers.

\paragraph{Extraction audits.}
From 350 original responses outside the requested format, we extracted 4,866 literal answer values, including 64 partial patterns, with verified source locations. A check of 60 credited answers extracted from prose, performed without consulting the reference answers, covered 42 languages and 13 model--condition groups. All were supported by the response and agreed with the scoring table. This audit verifies extraction; it does not assess cultural validity.

We checked 80 entries scored as incorrect or missing, drawn from 77 responses across all 11 models and five conditions. A fixed hash selected one incorrect or missing entry per model--condition pair; 25 additional entries covered response formats where extraction might miss an answer. Candidates were selected before checking the references and previously extracted answers. No complete answer had been missed in this sample. One Portuguese E response gave two letter positions not converted into a partial mask. This does not affect whole-answer scoring or coverage because the entry lacks a complete answer. Analyses of partial units would require systematic recovery. The audit does not establish the overall proportion of supplied answers recovered or assess their semantic validity.

\paragraph{Selection for recovery.}
We reviewed 381 answer-list responses that did not follow the requested structure. Applying the classification rules in order, we set aside 22 fully accounted prose outputs, three answer mappings recoverable by correcting syntax, three abbreviated mappings, two explicit declines and one explicit partial attempt. Of the remainder, 38 mentioning \texttt{final\_answer} or ``final answer field'' were selected for recovery; 28 literal ellipses were treated as nonanswers; and 284 residual summaries formed the second recovery group. Two grid ellipses bring the nonanswer total to 30. The reproducibility materials include the classification rules, exceptions and response-level decisions.

These rules select 322 responses: 171 contain some fully specified answers and 151 contain none. This selection includes field claims and completion summaries, as well as some inability statements, so it does not establish storage loss. A recovery response must come from the intended model, complete execution and contain one unambiguous answer map with the expected clue identifiers. Missing or wrong answers within an otherwise usable map are scored normally. The analysis uses 322 replacements (145 Opus 5.5, 127 Opus 4.7, 50 Fable 5.1).

One recovery JSON object contained a literal control character; decoding permitted that character without changing answer text. All replacement mappings preserve the source clues, reference keys and original identifiers. The archive records original and replacement response hashes, prompt hashes, model identifiers and retained timestamps. Retry selection depended on response availability and did not use reference agreement. Unequal attempts and the explicit-output prompt change preclude interpreting improvement as the isolated effect of either model ability or capture repair.

\paragraph{Crossing checks.}
Validation mapped all 5,057 reference entries and found agreement at all 7,236 physical intersections. Candidate mapping preserves whole units and right-to-left rules, counts intersections once, and excludes partial descriptions and wrong-length answers. Conditional consistency requires checkable crossings: 566 B records and 545 E records qualify, including 535 matched pairs. The paired increase is 18.7 points [14.3, 23.1]. A further sensitivity uses only intersections checkable in both conditions. For the eight models without recovery reruns, consistent/all crossings increase by 10.6 points [8.0, 13.3]. These selected comparisons supplement the full-denominator result.

\section{Alternative answers and analysis subsets}\label{app:populations}\label{app:alternatives}
The benchmark contains 53 puzzles, 5,057 puzzle-specific entries and 2,915 responses (278,135 entry observations). Reused riddles and crossing cells are not independent observations. Source annotations labelled category 3.1 identify answers accepted by human reviewers despite differing from the reference. We require consistent human score 3 (correct), no conflicting scores, unique source-ID matches, and clues matching after conservative Unicode/HTML/whitespace handling. This yields 1,984 clue--answer pairs covering 988 entries. Prompt-specific normalisation and deduplication leave 1,950 forms, including 60 reference matches. The remaining 1,890 forms cover 962 clues, 38 puzzles and 30 language labels, excluding Japanese. Original strings and evidence are retained without inferring synonyms or splitting explanations.

For normalised recorded non-reference forms $S_i$, written-unit sequence $u(a)$, required length $\ell_i$, and supplied hint positions/units $H_i,h_{ik}$, define
\[
S_i^{B}=\{a\in S_i:|u(a)|=\ell_i\},\qquad
S_i^{C}=\{a\in S_i^{B}:u(a)_k=h_{ik}\ \forall k\in H_i\}.
\]
Compatibility was checked against all 5,057 reference entries. Lengths exclude 91.7\% of the listed alternatives and lengths with hints exclude 96.5\%. Length leaves 156 forms and hints leave 66; no recorded form survives for 824 and 903 entries, respectively. A sensitivity filter removes possible lists/disjunctions, parentheticals, long explanations and format-control characters, leaving 1,584 forms; length/hint exclusions are 90.5\%/95.8\%. These checks do not validate meaning; removing source-withdrawn clues leaves the cohort unchanged.

The subset contains 42,328 entry observations across A/B/C/E and all 11 models. C--B gains are 9.6 [7.0, 12.4] for reference agreement and 8.4 [5.8, 10.9] for reference-or-alternative agreement; E--B gains are 4.6 [1.6, 7.4] and 3.0 [0.3, 5.4], respectively, in percentage points. We retain gains and losses in all paired comparisons. E's crossings are not supplied reference letters.
\begin{table}[tb]\centering\small\caption{Fractions show transitions from a listed alternative to the reference divided by all transitions to the reference after adding lengths. Full samples contain 962 clues per model; exclusions remove originally flagged pairs. Losses remain in the archived data.}\label{tab:historical-models}
\begin{tabular}{lrrr}\toprule Model & Full: alternative/gains & Excluded: alternative/gains & Retained clues\\\midrule
GPT-5.5 & 129/272 & 129/272 & 962 \\
GPT-6 Astra & 138/293 & 138/293 & 962 \\
GPT-6.1 Sol & 140/287 & 140/287 & 962 \\
Gemini 3.1 Pro & 56/163 & 56/163 & 962 \\
Gemini 3.7 Flash & 107/237 & 107/237 & 962 \\
Gemini 3.8 Flash & 112/284 & 112/284 & 962 \\
Gemini 4 Argon & 139/289 & 139/289 & 962 \\
Claude Fable 5.1 & 123/291 & 77/181 & 595 \\
Claude Opus 4.7 & 96/261 & 7/31 & 200 \\
Claude Opus 4.8 & 24/75 & 24/75 & 962 \\
Claude Opus 5.5 & 127/286 & 19/37 & 123 \\
\bottomrule\end{tabular}\end{table}

Two examples illustrate changes in wording and interpretation. For US R018, GPT-5.5 changes from \texttt{COCACOLA} to \texttt{COKE}. For US R022, ``Slavery in the South Under Another Name,'' it changes from the accepted alternative \texttt{SHARECROPPING} to \texttt{JIMCROWLAWS}. These name related but distinct institutions; this interpretation has not had independent specialist validation.

The English review has 37 forms across 19 clues (11 US, eight Canadian): 19 naming/wording differences, 11 scope differences, six possible changes in what the answer refers to across four clues, and one potential reference problem. Labels use references, accepted alternatives and conditions, without independent validation. Canadian R038's \texttt{SAINTCATHERINE} and reference \texttt{LASTECATHERINE} both have 14 units and survive the hints. This inventory example is not an observed model transition.

\section{Sensitivity and resampling}\label{app:sensitivity}
The 55 model pairs across five conditions and two samples (all responses, and pairs not selected for recovery) yield 550 comparisons, including puzzle wins, ties and losses. The 110 within-model contrasts (five contrasts, 11 models, two recovery-selection groups) use 10,000 language-cluster draws, seed 20261018, preserving membership and equal puzzle weights. Every interval agrees with the archived puzzle-bootstrap interval (seed 20261017) on whether it includes zero. Subsets with fewer than ten clusters are descriptive. Model comparisons and alternative-answer aggregates also use 10,000 language-cluster draws, seed 20261018; crossing contrasts use seed 2026101806. Alternative-answer aggregates average models within puzzle, then weight the 38 puzzles equally.

We retain the same original flag-exclusion cohorts after recovery. Fable's unflagged E--B comparison uses 27 puzzles, and Opus 5.5's uses only three. Such selected subsets cannot define a corrected general ranking. The archive includes original and recovery-integrated results for all 55 model--condition means and all replaced records.\begin{table}[htbp]\centering\small\caption{Original and recovery-integrated means for the three affected models (percent reference agreement). Each cell shows original $\rightarrow$ integrated; D is unchanged. The same 53 puzzles are retained.}\label{tab:recovery}
\begin{tabular}{lrrrr}\toprule Model & A & B & C & E\\\midrule
Claude Opus 5.5 & 28.3$\to$52.9 & 29.3$\to$76.6 & 31.4$\to$85.2 & 21.6$\to$82.6 \\
Claude Fable 5.1 & 46.6$\to$54.6 & 65.4$\to$77.8 & 73.4$\to$86.1 & 61.2$\to$84.1 \\
Claude Opus 4.7 & 26.5$\to$46.4 & 18.7$\to$65.8 & 23.3$\to$75.3 & 22.7$\to$58.8 \\
\bottomrule\end{tabular}\end{table}

{\emergencystretch=2em
The eight-model alternative-answer panel gives reference/\allowbreak reference-or-alternative gains of 21.1/7.4 points for B--A, 9.4/8.0 for C--B and 5.1/3.2 for E--B, using the same clues and puzzles as the primary panel. The clues and puzzles are held fixed; the model panel changes.\par}

\paragraph{Model omissions.}
We omit Opus 4.8, then each other model separately, retaining the same ten models across all five conditions in each comparison. Table~\ref{tab:model-omission} uses 10,000 language-cluster draws (seed 2026101806, also used for crossings); alternative-answer comparisons and regressions retain seed 20261018. Without Opus 4.8, E--B consistent crossings increase by 11.2 [8.9, 13.5] points while checkable coverage still falls. B--A reference and alternative-inclusive gains on this subset remain distinct (24.7 versus 9.2 points).
\begin{table}[htbp]\centering\small
\begin{tabular}{lccc}\toprule
Reference-agreement contrast & All 11 & Without Opus 4.8 & Omission range\\\midrule
Lengths: B--A & 16.9 [14.2, 20.0] & 18.5 [15.5, 21.9] & 16.2--18.5 \\
Hints: C--B & 8.4 [6.5, 10.1] & 8.8 [6.8, 10.7] & 7.8--8.8 \\
Crossword information: E--B & 2.8 [1.3, 4.3] & 4.2 [2.5, 5.8] & 2.2--4.2 \\
Answer list over grid: E--D & 26.6 [24.3, 28.6] & 26.8 [24.6, 28.6] & 22.2--29.1 \\
\bottomrule\end{tabular}
\caption{Model-omission sensitivity (percentage points). Brackets are pointwise 95\% language-cluster intervals. The last column spans mean effects when each model is omitted in turn; it is not an interval. All four intervals remain above zero in every omission. The same 53 puzzles are used throughout.}\label{tab:model-omission}
\end{table}

Excluding Opus 4.8 gives Spearman rank correlations (agreement between rankings) of 0.981--0.993 across the 53 language--country sets in A/B/C/E, and 0.984--0.996 across 42 language summaries. Removing Opus 4.8 leaves grid-task rankings unchanged because it scores zero on every puzzle. Individual positions can still change substantially: Portuguese/Brazil moves from 17th to 28th in E. Writing-system mean order is unchanged by this exclusion; other model omissions reverse small abjad--abugida differences in B and E. These comparisons do not isolate script from content or construction. Across all model omissions, regression $R^2$ ranges from 0.017 to 0.091 and every slope interval includes zero. Detailed ranks, group contrasts and coefficients accompany the analysis code.

\section{Puzzle-characteristic regression}\label{app:regression}
The outcome is each puzzle's mean E--B reference-agreement difference across the fixed model panel, in percentage points. Predictors are centred over 53 puzzles and scaled as in Table~\ref{tab:puzzle-regression}. Crossing density is the fraction of an entry's positions shared with another entry, averaged across entries. The intercept is the panel's mean gain. The fit has 53 observations; repeated models and clues do not increase this count. Language-cluster resampling retains country variants, using seed 20261018 for 10,000 draws shared across panels.

\begin{table}[!htbp]\centering
\small
\setlength{\tabcolsep}{4pt}\begin{tabular}{lccc}
\toprule
Puzzle characteristic & All 11 & Fixed eight & Without Opus 4.8 \\
\midrule
Ten more clues & -1.00 [-4.32, 1.65] & 0.33 [-3.62, 3.56]  & -0.77 [-4.47, 2.22] \\
One more answer unit & 0.43 [-0.87, 1.57] & 0.98 [-0.83, 2.50]  & 0.25 [-1.16, 1.57] \\
0.10 more crossing density & 2.16 [-3.78, 7.22] & 5.95 [-2.18, 12.88]  & 1.57 [-5.13, 7.41] \\
\midrule
$R^2$ & 0.032 & 0.038  & 0.017 \\
\bottomrule
\end{tabular}
\caption{Conditional associations with the puzzle-level E--B reference-agreement gain (percentage points). Each fit uses 53 equally weighted puzzles and all three predictors. Brackets give pointwise 95\% intervals from 10,000 language-cluster bootstrap draws. Fixed eight excludes the three configurations selected for recovery. Answer length and crossing density are strongly correlated; these are descriptive associations, not causal effects.}\label{tab:puzzle-regression}
\end{table}

Clue counts range from 72 to 100; 34 puzzles have 99--100 clues. Variance-inflation factors, which measure overlap between predictors, are 1.51, 5.05 and 5.41 for size, length and crossing density. Leave-one-language-out fits and reduced specifications check sensitivity to individual language groups and to the strong length--crossing correlation. Inputs, coefficients, bootstrap draws and influence checks accompany the code. Intervals describe variation across puzzles and language groups, not repeated-generation uncertainty or the effects of selective recovery.

\section{Scores and answer coverage}\label{app:scores}
\begin{figure}[!htbp]\centering
\includegraphics[width=0.92\textwidth]{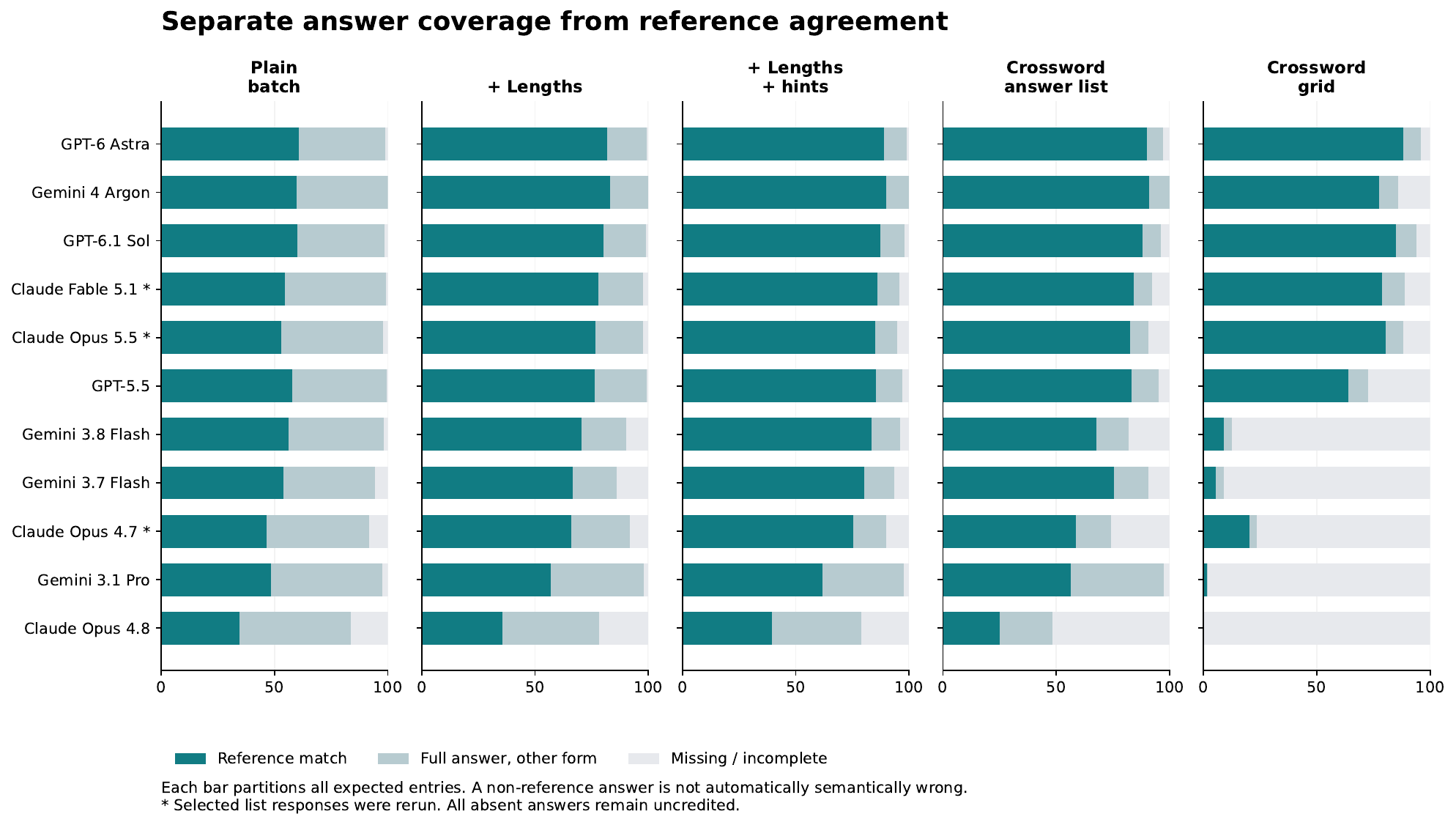}
\caption{Reference matches, fully specified non-reference answers, and incomplete or missing answers, using all expected clues as the denominator. Non-reference answers may be semantically acceptable. Selected list records use recovery attempts, and missing answers receive no credit.}\label{fig:coverage}
\end{figure}

Table~\ref{tab:results} retains all expected clues; D requires correct placement. Figure~\ref{fig:coverage} separates delivered answers from missing or incomplete material.
\begin{table}[!htbp]\centering\small
\caption{Reference agreement (percent), weighting 53 puzzles equally and retaining missing answers in the denominator. D requires correct grid placement. Selected answer-list records for Opus 5.5, Fable 5.1 and Opus 4.7 use recovery attempts with explicit inline-output instructions.}\label{tab:results}
\begin{tabular}{lrrrrr}\toprule Model & A & B & C & D & E\\\midrule
GPT-6.1 Sol & 60.2 & 80.2 & 87.4 & 85.0 & 88.1 \\
GPT-6 Astra & 60.7 & 81.7 & 89.0 & 88.3 & 89.9 \\
GPT-5.5 & 57.9 & 76.3 & 85.3 & 64.0 & 83.2 \\
Gemini 4 Argon & 59.8 & 83.1 & 90.0 & 77.4 & 91.1 \\
Gemini 3.8 Flash & 56.1 & 70.6 & 83.5 & 9.0 & 67.6 \\
Gemini 3.7 Flash & 53.9 & 66.5 & 80.3 & 5.4 & 75.6 \\
Gemini 3.1 Pro & 48.3 & 57.0 & 61.9 & 1.5 & 56.5 \\
Claude Fable 5.1 & 54.6 & 77.8 & 86.1 & 78.7 & 84.1 \\
Claude Opus 5.5 & 52.9 & 76.6 & 85.2 & 80.4 & 82.6 \\
Claude Opus 4.7 & 46.4 & 65.8 & 75.3 & 20.2 & 58.8 \\
Claude Opus 4.8 & 34.5 & 35.5 & 39.4 & 0.0 & 24.9 \\
\bottomrule\end{tabular}\end{table}

\clearpage

\begin{figure}[p]\centering
\includegraphics[width=\textwidth]{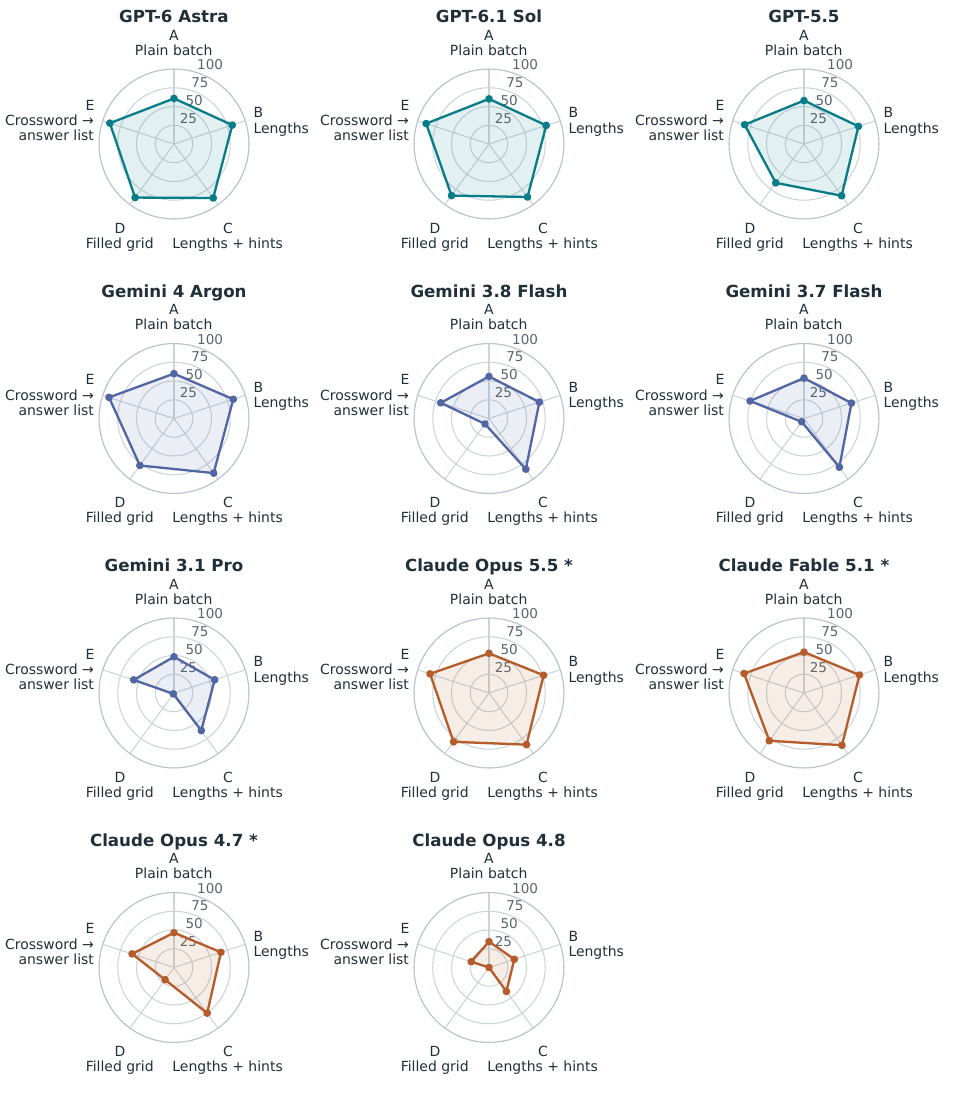}
\caption{Model profiles across the five task formats, using reference agreement averaged equally over 53 puzzles. All panels use the same 0--100\% scale and A--B--C--D--E axis order. D also requires correct placement; missing answers remain in the denominator. Corresponding spokes show each format's score; polygon area has no performance interpretation. Uncertainty intervals are omitted. Asterisks mark models with selected list responses replaced by recovery attempts under revised output instructions. Opus 4.8's zero in D reflects noncompletion under this protocol.}\label{fig:model-radars}
\end{figure}
\clearpage

\section{Task prompts}\label{app:prompts}
The excerpts below reproduce saved task instructions, with line breaks adjusted. They are not reconstructions of the full provider messages or system instructions. Each prompt also supplies the puzzle's language and regional context, language-specific spelling and cell-unit rules, and the complete clue set. The archive contains all 265 original task prompts and 188 distinct recovery prompts, with checksums and a mapping to all 2,915 retained responses. Full texts preserve the supplied grids, lengths, hints and output requirements.

\paragraph{Shared answer-list instructions (A, B, C and E).}
\begin{quote}\small
Choose one answer per clue using the clue and any supplied constraints. Do not combine alternatives or append explanations. Duplicate answers are allowed.

Return answers in the same spelling and cell-unit representation required for this puzzle's crossword answers, even when no grid is supplied.

In every condition, answer strings use normal logical spelling order. Do not reverse a right-to-left answer to imitate its physical grid placement.
\end{quote}
These prompts explicitly prohibit internet access, code execution and other external tools. They apply the source puzzle's answer conventions, including specified punctuation removal and script conversion, where required.

\paragraph{Condition-specific instructions.}
\begin{description}\small
\item[A:] ``No answer lengths, letter hints or crossword layout are supplied. Solve the clues using their text and the supplied language and regional context.''
\item[B:] ``Each clue includes length\_units: the required number of answer units after applying the supplied formatting conventions. Match this length. No letter hints or crossword layout are supplied.''
\item[C:] ``Each clue includes length\_units and known\_units. Positions are one-based positions in the formatted answer in normal logical spelling order. Every supplied unit is correct. Match the length and all supplied units. No crossword layout is supplied.''
\item[D, excerpt:] ``Use the grid structure and intersecting letters to constrain your answers. Determine each answer's required length from the grid.''
\item[E:] ``The supplied grid, directions and starting coordinates define a crossword. Each clue also includes its required length\_units. Solve the clues jointly, using intersecting cells to constrain the answers. Entries sharing a cell must agree on that cell. Return the answer to each clue using its ID in normal logical spelling order; do not return the grid.''
\end{description}

\paragraph{Required outputs.}
A/B/C/E request a JSON object with \texttt{puzzle\_code} and an \texttt{answers} mapping from clue identifiers to answer strings:
\begin{quote}\small
Include every clue ID exactly once, in the supplied order. If you cannot give any answer for a clue, use an empty string. You may use ``?'' for unknown units within a partially known answer; retain units you know.

Do not include explanations, alternative answers, model identification, Markdown fences, additional fields or attached files.
\end{quote}
D instead requests \texttt{puzzle\_code} and \texttt{rows}: 50 strings of 50 cells, read top to bottom and physically left to right, including for right-to-left languages. It requires preservation of blocks and known cells, replacement of unknown cells with \texttt{?}, and no answer list. Its cell-unit wording varies by puzzle: 33 prompts specify one Unicode code point per cell; 20 specify one extended grapheme cluster, which can contain multiple code points. The Russian D prompt omits the internet prohibition present in the other 52; recorded tool settings still disable web access.

\paragraph{Recovery addition (A/B/C/E only).}
For all 322 replacements, the recovery prompt preserves the original task prompt as an exact prefix and appends the following wording. Here, \texttt{N} stands for the supplied clue count; the full archive retains the number used in each task.
\begin{quote}\small
\textbf{OUTPUT DELIVERY --- INLINE TEXT ONLY}

Write the complete required JSON object, containing puzzle\_code and answers for all \texttt{N} clue IDs, directly in the text of your response. Do not create, save, attach or link to any file, document, artifact or downloadable JSON. Do not say that the answers were supplied elsewhere, and do not substitute a summary of your work for the answers. The answer strings themselves must be present in the response text.

If the surrounding response format requires separate solution and final\_answer fields, put the complete required answer JSON as text in the solution field as well as in final\_answer. A summary or a reference from one field to another is not sufficient. Do not omit any clue ID; use an empty string for an unanswered clue as instructed above. Before finishing, ensure that the response text actually contains every clue ID and its answer.
\end{quote}
This addition changes output-delivery instructions, not the clues or constraints. Its selective use is assessed by the recovery sensitivities in Appendix~\ref{app:sensitivity}.

\end{document}